\pdfoutput=1

\documentclass[11pt]{article}

\usepackage[]{acl}

\usepackage{times}
\usepackage{latexsym}

\usepackage[T1]{fontenc}

\usepackage[utf8]{inputenc}

\usepackage{microtype}

\usepackage{inconsolata}

\usepackage[final]{changes}
\usepackage{graphicx}
\usepackage{bm}
\usepackage{amsmath}
\usepackage{booktabs}
\usepackage{multirow}
\usepackage{amsfonts}
\usepackage{tabularx}
\usepackage{array}
\usepackage{bbding}
\usepackage{makecell}
\newcolumntype{Y}{>{\centering\arraybackslash}X}  
\newcolumntype{Z}{>{\raggedright\arraybackslash}X}  

\title{CauESC: A Causal Aware Model for Emotional Support Conversation}

\author{
    Wei Chen\textsuperscript{\rm 1}, 
    Hengxu Lin\textsuperscript{\rm 2},
    Qun Zhang\textsuperscript{\rm 1},  
    Xiaojin Zhang\textsuperscript{\rm 1},   \\
    \textbf{
    Xiang Bai\textsuperscript{\rm 1},  
    Xuanjing Huang\textsuperscript{\rm 2},
    Zhongyu Wei\textsuperscript{\rm 2}\thanks{~~Corresponding Author.}}
    \\ [.3cm]
    \textsuperscript{\rm 1}{ Huazhong University of Science and Technology} \\
    \textsuperscript{\rm 2}{ Fudan University} \\
    \{lemuria\_chen, qunzhang, xiaojinzhang, xbai\}@hust.edu.cn \\
    linhx21@m.fudan.edu.cn,  
    \{xjhuang, zywei\}@fudan.edu.cn 
}

\begin{document}
\maketitle
\begin{abstract}
Emotional Support Conversation aims at reducing the seeker's emotional distress \replaced{through}{with} supportive \replaced{response}{responses}. Existing approaches have two limitations: (1) They ignore the emotion causes of the distress, which is important for fine-grained emotion understanding; (2) They focus on the seeker's own mental state rather than the emotional dynamics during interaction between speakers. To address these issues, we propose a novel framework CauESC, which firstly recognizes the emotion causes of the distress, as well as the emotion effects triggered by the causes, and then understands each strategy of verbal grooming independently and integrates them skillfully. Experimental results on the benchmark dataset demonstrate the effectiveness of our approach and show the benefits of emotion understanding from cause to effect and independent-integrated strategy modeling. 

\end{abstract}

\section{Introduction}

In recent years, dialogue systems have experienced a booming with a variety of neural based models \cite{serban2016building, xing2018hierarchical, zhang2019dialogpt, roller-etal-2021-recipes, chen2022contextual}. Despite the potential benefits, these systems also face several challenges, including the tendency to generate unhelpful or even harmful content, lack of empathy, and difficulty in resonating with users. Such open domain dialogue systems can pose risks for individuals experiencing emotional anxiety. \added{Concurrently, numerous researchers have undertaken the curation of dialogues enriched with emotion annotations \cite{li2017dailydialog, sharma-etal-2020-computational, wang-torres-2022-helpful, liu-etal-2021-towards}.}

\deleted{To enhance the empathy of the dialogue system, Emotional Support Conversation \cite{liu-etal-2021-towards} has attracted much attention.} \added{
In contrast to the conventional approach of directly identifying anxious users \cite{ijcai2020p760}, measures aimed at alleviating anxiety through the enhancement of empathy in Emotional Support Conversation \cite{liu-etal-2021-towards} attract considerable attention.} Emotional support (ES) aims at reducing individuals’ distress and helping them understand and work through the challenges that they face \cite{burleson2003emotional, langford1997social, heaney2008social}.

\begin{figure}
  \centering
  \includegraphics[width=0.5\textwidth]{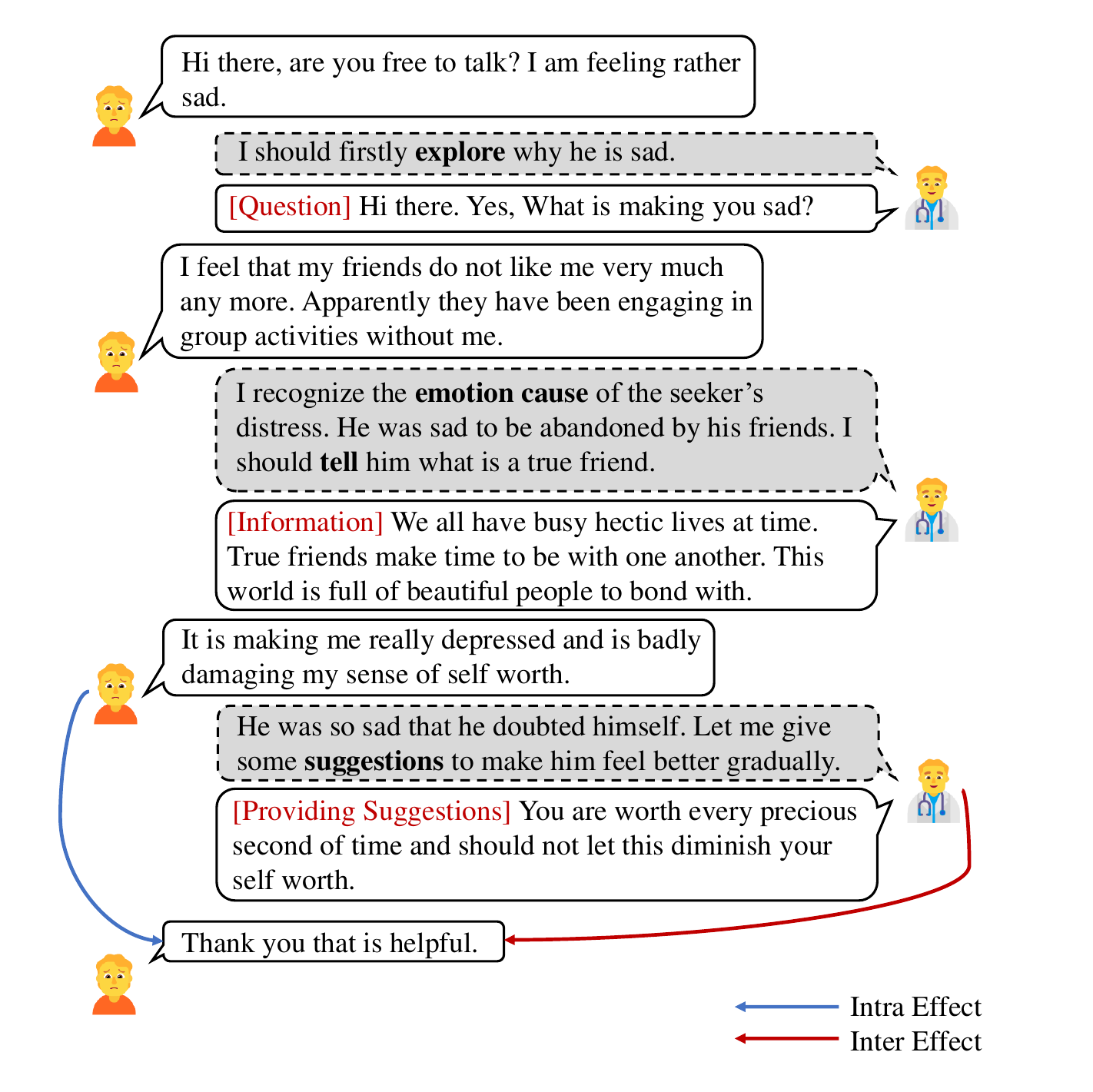}
  \caption{An Emotional Support Conversation Example, which intra and inter effects gradually change the emotional state of the seeker.}
  \label{fig:esconv}
\end{figure}

Existing work has attempted to endow the model with empathy from the perspective of fine-grained user state modeling \cite{tu-etal-2022-misc} and hierarchical psychological relation modeling \cite{DBLP:conf/ijcai/00080XXSL22}.Some researchers have also explored manual annotation and training data augmentation through policy annotations \cite{cheng-etal-2022-improving}. The others also approach empathy from a pragmatic perspective \cite{kim2021perspectivetaking}. However, despite these efforts,  there are still some limitations including a lack of consideration for the underlying \textit{emotion cause} of the seeker's distress and \textit{emotional dynamics} which refers to emotion effects during the interaction between speakers. As shown in Figure \ref{fig:esconv}, emotional dynamics can be divided into Intra Effect and Inter Effect, which represent the negative effect of the seeker's distress and the supportive effect of the supporter on the user, respectively.

\deleted{To overcome these issues, we propose \textbf{CauESC}, i.e.,  a \textbf{Cau}sal Aware Model for \textbf{E}motional \textbf{S}upport \textbf{C}onversation. CauESC leverages an emotion cause detector to recognize the cause of the seeker's distress and uses a special attention mechanism to focus on these causes. Additionally, considering emotional dynamics, CauESC incorporates COMET \cite{bosselut-etal-2019-comet} to capture the intra and inter effects during the conversation and devises an attention mechanism to infer the effects triggered by the emotional causes. To further improve the flexibility of generation, CauESC adopts an independent-integrated strategy modeling approach with an innovative design of independent-integrated strategy executors. The unique design not only helps to understand the semantic information of each strategy independently, but also integrates multiple strategies to guide the response generation.}

\added{To address these issues, we propose \textbf{CauESC}, i.e.,  a \textbf{Cau}sal Aware Model for \textbf{E}motional \textbf{S}upport \textbf{C}onversation. CauESC utilizes an emotion cause detector and a specialized attention mechanism to recognize and focus on the causes of the seeker's distress. Incorporating COMET \cite{bosselut-etal-2019-comet} to account for intra and inter effects, CauESC employs an attention mechanism to infer effects triggered by emotional causes. Enhancing generation flexibility, CauESC adopts an independent-integrated strategy modeling approach with an innovative decoder design, facilitating both independent strategy comprehension and integrated strategy use.}


\deleted{To evaluate our model, we conduct extensive experiments on ESConv benchmark \cite{liu-etal-2021-towards} and compare with 11 state-of-the-art dialogue systems including MISC \cite{tu-etal-2022-misc}, TransESC \cite{zhao2023transesc}, MultiESC \cite{cheng-etal-2022-improving} and ChatGPT \cite{OpenAI2022}. Based on both automatic evaluations and human evaluations, we demonstrate that the responses generated by our model CauESC are more supportive, especially Identification and Comforting. Besides, the case study sheds light on how emotion causes and effects could contribute to the generation of supportive responses.}

\added{For evaluation, we conduct extensive experiments on the ESConv benchmark \cite{liu-etal-2021-towards}, comparing with 9 state-of-the-art dialogue systems, including MISC \cite{tu-etal-2022-misc} and ChatGPT \cite{OpenAI2022}.Through both automatic and human evaluations, we demonstrate that CauESC generates more supportive responses, particularly in Identification and Comforting.}

\deleted{In brief, our contributions \replaced{include}{are as follows}: (1) We present \replaced{an}{a} emotional support dialog system CauESC, which incorporates emotion cause and effect into emotional support conversation. To the best of our knowledge, this is the first work that investigates emotion cause and effect in emotional support conversation.  (3) We propose the independent-integrated strategy modeling approach to independently understand each strategy and integratedly adopt multiple strategies to guide response generation. (4) We conduct experiments on ESConv dataset, and demonstrate CauESC achieves the state-of-the-art performance in terms of both strategy selection and response generation. }
\added{In brief, our contributions include (1) presenting an emotional support dialog system CauESC, the first model investigating emotion \deleted{cause and effect} \added{ causes and intra/internal effects of emotions} in emotional support conversation, (2) devising an attention mechanism for reasoning about effects triggered by emotion causes, mimicking human cognitive processes, (3) proposing an independent-integrated strategy modeling approach for understanding and using multiple strategies in response generation, and (4) achieving state-of-the-art performance on strategy selection and response generation according to experiments on the ESConv dataset.}

\section{Related Work}
\subsection{Emotional Support Conversation Model}
In general, emotional support conversation models can be divided into two categories, including explicitly model psychological factors at the cognitive level using graph neural networks and generating supportive responses by utilizing support strategies. For instance, \citet{DBLP:conf/ijcai/00080XXSL22} propose a global-to-local hierarchical graph network to establish hierarchical relationships among psychological factors for generation. For strategy-driven response generation, \citet{liu-etal-2021-towards} append a special token to represent each strategy before the response to make a generation. \citet{tu-etal-2022-misc} design a mixed strategy-aware model integrating COMET \cite{bosselut-etal-2019-comet}, a pre-trained commonsense language model, to respond skillfully. However, they ignore the emotion causes of the distress and the emotional dynamics \cite{poria2019emotion}. Recent work also includes K-ESConv~\cite{chen2023k}, which uses GPT-2 combined with trainable prompting methods to model emotional dialogue.

\subsection{Commonsense Reasoning}
Existing work on generation of novel commonsense has used ATOMIC \cite{sap2019atomic} as underlying KBs. Specifically, they use LSTM encoder-decoder models to generate commonsense about social situations. \citet{bosselut-etal-2019-comet} use transformers and investigate the effect of using pre-trained language representations \cite{radford2018improving} to initialize them. \citet{turcan-etal-2021-multi} demonstrate that neural networks can predict the probable causes and effects of previously unseen events with the help of the commonsense provided by ATOMIC. Inspired by the above work, we use COMET, a commonsense reasoning model trained over ATOMIC, to reason the commonsense about the emotion effects. 

\subsection{Emotion Cause Extraction}
Emotion cause extraction (ECE), aims at exploring the reason for emotion change and what causes a certain emotion. \citet{gui-etal-2016-event} propose the first open dataset for ECE. \citet{xia-ding-2019-emotion} reform ECE into emotion-cause pair extraction task. Similar to ECE, \citet{poria2021recognizing} firstly introduce the task of recognizing emotion cause in conversations. \citet{gao-etal-2021-improving-empathetic} firstly investigate emotion cause in empathetic response generation, resulting in more empathetic responses. Inspired by the above work, we recognize emotion causes at utterance level in emotional support conversation to generate more supportive responses. Identifying the reasons behind emotions is not only crucial for emotional support dialogue systems but also plays a significant role in healthcare dialogue systems~\cite{chen2023benchmark,zhong2022hierarchical,chen2023dxformer,lopez2023case}.

\section{Problem Formulation}
Emotional Support Conversation requires the model to firstly select a strategy and then provide a supportive response with the strategy to reduce the seeker's distress. The input has three components: (1) a situation $\bm q$ which describes the seeker’s problem in free form text; (2) a multi-turn emotional support conversation consisting of $M$ utterances $\bm{C}=\left(\bm{u}_1, \bm{u}_2, \ldots, \bm{u}_M\right)$, where the speaker of the last turn is the seeker; (3) a strategy history for each turn of the supporter $\bm{S}=\left(s_1, s_2, \ldots, s_J\right)$, where $J$ denotes the number of turns by the supporter. Our model aims to select a strategy and generate the supportive response $\bm{y}$ with the given $\bm{q}$, $\bm{C}$ and $\bm{S}$. Consequently, the target becomes to estimate the probability distribution $p\left(\bm{y}|\bm{q},\bm{C},\bm{S}\right)$.

\section{Approach}
\begin{figure*}[htb]
\centering
\includegraphics[width=1\textwidth]{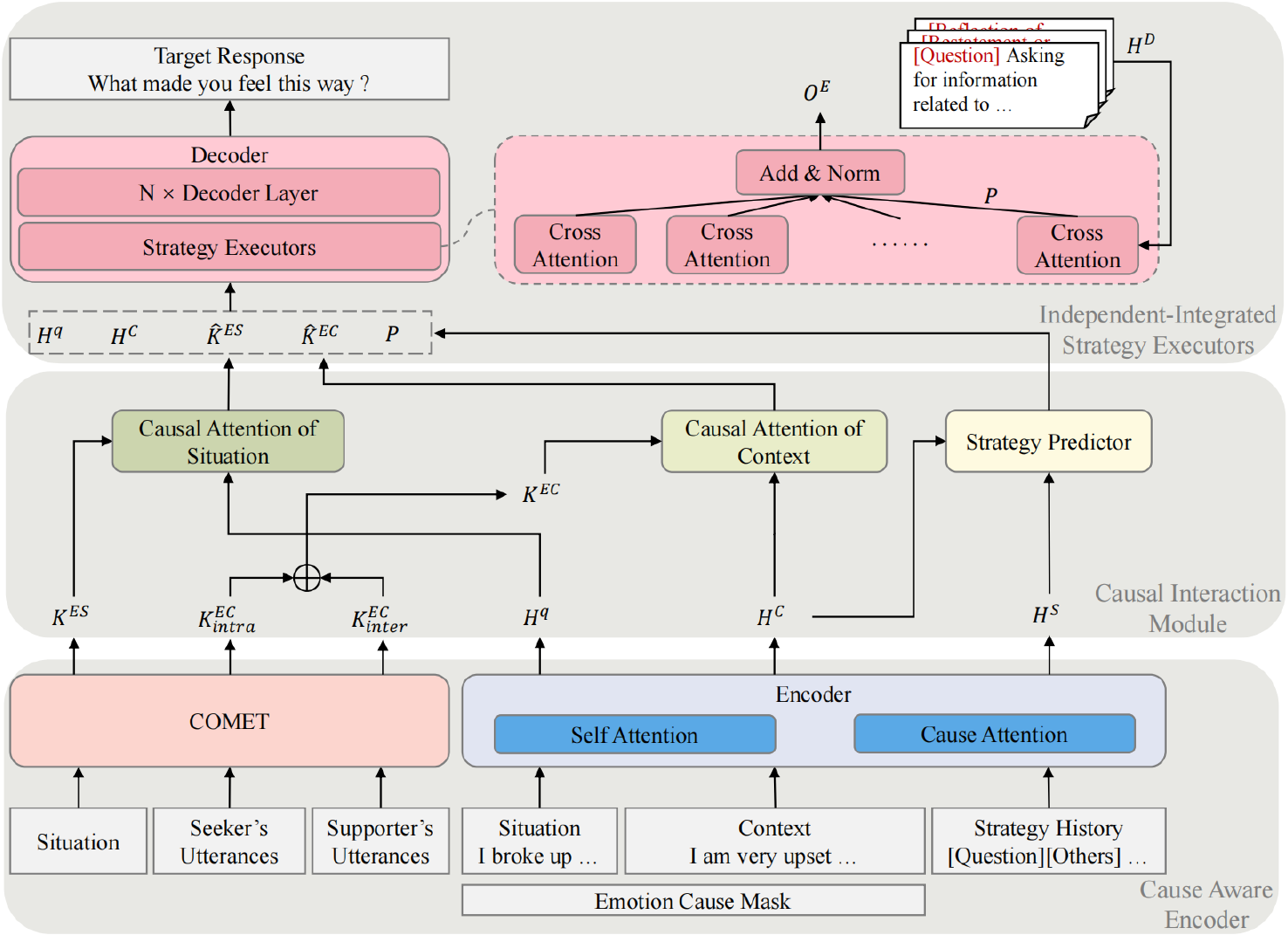}
\caption{The overview of the proposed CauESC which consists of a cause aware encoder, a causal interaction module, and independent-integrated strategy executors. Gray dotted curves indicate the detail of the corresponding module.}
\label{fig:cauesc}
\end{figure*}

The overview of our approach is shown in Figure \ref{fig:cauesc}. Based on blenderbot-small \cite{roller-etal-2021-recipes}, our model CauESC consists \added{of} three main components: (1) a cause aware encoder; (2) a causal interaction module; and (3) independent-integrated strategy executors.

\subsection{Cause Aware Encoder}




The cause aware encoder is used to recognize the emotion causes of the seeker's distress and capture the emotion effects in the conversation. It can be decomposed into two tasks: emotion cause encoding and emotion effect acquisition.

\paragraph{Emotion Cause Encoding}
{To identify emotional causes, we use an existing emotion cause detector \cite{poria2021recognizing} trained on an open-domain emotional dialogue dataset called RECCON \cite{poria2021recognizing}. The detector determines utterances in a context that cause emotions in the target utterance. Input includes context, target utterance, and emotion label, with binary labels for each context utterance. The detector identifies sentences causing the seeker's emotion, using the seeker's last utterance as the target, and corresponding sentiment label. To focus on emotion causes, we introduce Cause Attention, leveraging emotion causes for encoding. Unlike self-attention, Cause Attention emphasizes utterances with emotion causes.Given the context hidden state $\bm{H}$ of length $N$ and the emotion cause mask $\bm{E} = \left(e_1, e_2, \ldots, e_N\right)$, where the mask values are in the range $\{0,1\}$, the cause attention distribution $a_{ij}$ is computed as follows:
\begin{equation}
    a_{ij} = \frac{e_i \odot \mathrm{exp}\left(q_i k_j/\sqrt{d}\right)}{\sum_{n=1}^{N} e_n \odot \mathrm{exp}\left(q_i k_n/\sqrt{d}\right)}
\end{equation}
\added{where $q_i$ is the query vector for the $i$-th position, $k_j$ is the key vector for the $j$-th position, and $d$ is the dimension of the model.}

The cause attention and self attention are parallel, which will output $\bm{A}^c$ and $\bm{A}^s$ respectively:
\added{
\begin{equation}
    \begin{aligned}
        \bm{A}^s = & \text{SELF-ATT}\left(\bm{H}\right) \\
        \bm{A}^e = & \text{CAUSE-ATT}\left(\bm{H},\bm{E}\right)
    \end{aligned}
\end{equation}
where $\mathrm{SELF\text{-}ATT(\cdot)}$ stands for self-attention calculation and $\mathrm{CAUSE\text{-}ATT(\cdot)}$ stands for cause-attention calculation.}

The outputs of both will pass through a fusion layer which consists of a linear layer coupled with a ReLU activation function:
\begin{equation}
    \begin{aligned}
        \mu = & \mathrm{ReLU} \left (\bm{W}_\mu \left [\bm{A}^c;\bm{A}^s \right ] + \bm{b}_\mu \right ) \\
        \bm{F} = & \mu \cdot \bm{A}^s + \left(1-\mu\right) \cdot \bm{A}^c
    \end{aligned}
\end{equation}
where $\bm{W}_\mu$ and $\bm{b}_\mu$ are learnable parameters. Next, $\bm{F}$ is fed to a connected feed-forward network with residual connections and layer normalization to obtain the cause aware context.

\added{For context $\bm{C}$, the comforting strategies used by the speaker and the comforter are separated by the special participle $\texttt{[SEP]}$ to add speaker information and comforting strategy information. The context $\bm{C}$ will be preprocessed as follows:
\begin{equation}
        \begin{aligned}
            \bm{C} = & ([\texttt{USER}], u_1, [\texttt{SEP}], [\texttt{Question}], \\
            & u_2, \ldots, u_M)
        \end{aligned}
\end{equation}}

To comprehensively recognize the emotion causes, the cause aware encoder (CAE) encodes the situation and the context:
\begin{equation}
    \begin{aligned}
        \bm{H}^q = & \mathrm{CAE}\left(\bm{q}, \bm{E}^q\right) \\
        \bm{H}^C = & \mathrm{CAE}\left(\bm{C}, \bm{E}^C\right) \\
        \bm{H}^S = & \mathrm{CAE}\left(\bm{S}\right)
    \end{aligned}
\end{equation}
\added{where $\bm{E}^q$ and $\bm{E}^C$ represent the emotion cause mask for the situation and the context, respectively.$\bm{H}^q$, $\bm{H}^C$ and $\bm{H}^S$ represent the vector representations of situation, context and strategy history after cause-aware encoding, respectively. }\deleted{For strategy selection, the strategy history $\bm{S}$ is encoded as $\bm{H}^S$.} Since there is no emotion cause for the strategy history $\bm{S}$, the encoder ignores the cause attention,, which is consistent with a vanilla Transformer encoder.}

\paragraph{Emotion Effect Acquisition}
{To obtain the emotion effects based on emotional dynamics, we adopt the generative commonsense transformer model COMET \cite{bosselut-etal-2019-comet}. Specifically, we explore six relation types which are all categorized as “effects” based on their causal relations. \texttt{xReact}, \texttt{xEffect} and \texttt{xWant} are viewed as the intra relations, while \texttt{oReact}, \texttt{oEffect} and \texttt{oWant} are explained for inter relations. The definition of COMET relations is given in Appendix \ref{sec:comet}.

Given an utterance $\bm u_i$ (which is referred as \replaced{an}{a} event) and the selected relation type, COMET would generate descriptions of “then” with the format of if-then reasoning. We concatenate $\bm u_i$ and a relation with mask tokens such as $ \left(\bm{u}_i \ \texttt{[MASK]} \ \texttt{oReact}\right) $ to construct the input of COMET. Following \citet{ghosal-etal-2020-cosmic}, the hidden states from the last encoder layer of COMET are taken as the representation of effects.

To comprehensively capture emotion effects, the situation, the seekers' utterances and the supporter's utterances are fed into COMET to obtain emotion effects:
\begin{equation}
    \begin{aligned}
        \bm K^{ES} = & \bigcup_{rel \in \bm R_{intra}} \mathrm{COMET}\left(rel, \bm s\right)    \\
        \bm K^{EC}_{intra} = & \bigcup_{rel \in \bm U_{sek}}\bigcup_{rel \in \bm R_{intra}} \mathrm{COMET}\left(rel, \bm s\right)  \\
        \bm K^{EC}_{inter} = & \bigcup_{rel \in \bm U_{sup}}\bigcup_{rel \in \bm R_{inter}} \mathrm{COMET}\left(rel, \bm s\right)
    \end{aligned}
\end{equation}
Since the situation is generated by the seeker, we capture the intra effect of situation $\bm K^{ES}$ and context $\bm K^{EC}_{intra}$ from the situation and seeker's utterances based on intra relations $R_{intra}$, respectively. However, for the supporter's utterances, we obtain the inter effect of context $\bm K^{EC}_{inter}$ based on inter relations $R_{inter}$.}

\subsection{Causal Interaction Module}
The causal interaction module is responsible for reasoning to obtain the emotion effects triggered by the emotion causes, thus developing a fine-grained emotion understanding of the seeker.

Firstly, to comprehensively consider the emotion effects, we concatenate the aforementioned emotion effects, $\bm K^{EC}_{intra}$ and $\bm K^{EC}_{inter}$, to form a comprehensive effect representation $\bm K^{EC}$. \deleted{To infer the related emotion effects triggered by the emotion causes, }We creatively take attention method based on the cause aware context:
\begin{equation}
    \begin{aligned}
    \bm{\hat{K}}^{EC} = & \text{CROSS-ATT} \left (\bm K^{EC}, \bm H^{C}\right) \\
    \bm{\hat{K}}^{ES} = & \text{CROSS-ATT} \left (\bm K^{ES}, \bm H^{q}\right) 
    \end{aligned}
\end{equation}
where $\text{CROSS-ATT}$ represents a combination of cross attention, residual connection, and layer normalization. \deleted{Similarly, we could transform $\bm K^{ES}$ to $\bm{\hat{K}}^{ES}$ following the same method as $\bm K^{EC}$ to $\bm{\hat{K}}^{EC}$.}

Moreover, considering the strategy history is important for strategy selection, we integrate the strategy history $\bm{H}^S$ and the context representation $\bm{H}^C$ through mean-pooling. Then, the two representations are concatenated to form a query vector $h \in \mathbb{R}^{2d}$:
\begin{equation}
    \begin{aligned}
        s = & \text{Mean-pooling}\left(\bm{H}^S\right) \\
        c = & \text{Mean-pooling}\left(\bm{H}^C\right) \\
        h = & s \oplus c
    \end{aligned}
\end{equation}

Finally, we creatively learn from the idea of Key-Value Memory Network \cite{miller-etal-2016-key} to estimate the strategy distribution for strategy modeling. Specifically, we assign a key vector $k_i \in \mathbb{R}^{2d}$ as each strategy's latent vector. We assume that if the query vector is similar to a certain key vector, the probability of selecting the strategy is higher. Then, the query vector is used to address the key vector by performing a dot product followed by a softmax function. Thus, we have the strategy distribution as follows:
\begin{equation}
    p_i = \frac{e^{h^\top k_i}}{\sum_{j=1}^n e^{h^\top k_j}}
\end{equation}
where $n$ represents the number of strategies.

\subsection{Independent-Integrated Strategy Executors}
Existing work \cite{liu-etal-2021-towards, tu-etal-2022-misc} assumes that models implicitly learn how to respond supportively. However, without any additional inductive bias, a single decoder learning to respond to all strategies \added{loses in interpretability in the generation process and generates generic responses.}\deleted{will not only lose interpretability in the generation process, but will also generate more generic responses.}

To address this issue, the independent-integrated strategy executors are designed to understand the semantic information of each strategy independently, and combines multiple strategies to guide the generation from cause to effect. Specifically, we introduce strategy executors under the vanilla decoder which consist of independently parameterized executors. \added{Each executor is a cross-attention module responsible for understanding a specific strategy independently.}\deleted{Each executor is essentially a cross attention module that is responsible for understanding a specific strategy independently.}

Firstly, To consider the emotion causes and effects, we concatenate the cause-aware context and the emotion effects to generate a comprehensive causal representation:
\begin{equation}
    \bm X = \bm H^q \oplus \bm H^C \oplus \bm{\hat{K}}^{EC} \oplus \bm {\hat{K}}^{ES}
\end{equation}

Besides, to enrich the semantic information of the strategies, we introduce a speacial Strategy Reference whose key is a strategy and value is the corresponding description. The strategy names and their description are listed in Appendix \ref{sec:strategy_reference}. As with the strategy history, The strategy description $\bm D$ is also encoded to $\bm H^D$ without cause attention.

After that, we allow the executor to consider both the comprehensive causal information and the semantic information of the strategy:
\begin{equation}
    \bm O^E_i = \text{CROSS-ATT}_i\left(\bm O, \left[\bm X;\bm H^{D}_i\right]\right)
\end{equation}
where $\text{CROSS-ATT}_i$ refers to the $i$-th cross attention module (i.e., executor) and $\bm O$ represents the hidden state of the decoder.

Finally, to integrate the various strategies for generation, we combine the outputs from the executors by taking a weighted average with residual connections and layer normalization. The weights are determined by the strategy distribution. In addition, the decoder attends to the comprehensive causal information according to the integrated strategy output:
\begin{equation}
    \begin{aligned}
        \bm Z^E = & \sum_{i=1}^{n} p_i \bm O^E_i \\
        \bm O^E = & \text{LayerNorm}\left(\bm O + \bm Z^E\right) \\
        \bm O^\prime = & \text{DEC}\left(\bm O^E, \bm X\right)
    \end{aligned}
\end{equation}
where $\text{DEC}$ represents the decoder. Conceptually, each executor focuses on a specific strategy, while the decoder collects the strategy suggestions generated by executors to guide the response generation.

Based on \replaced{blenderbot-small}{blenderbor-small} \cite{roller-etal-2021-recipes}, we jointly train the model to select the strategy and generate the response. Specifically, given the target response $\bm y$ with a length of $\lvert \bm y \rvert$ and the target strategy $s_y$, the loss function is calculated as:
\begin{equation}
    \begin{aligned}
        \mathcal L_s = & - \mathrm{log}\left(p\left(s_y | \bm C,\bm S\right)\right) \\
        \mathcal L_r = & - \sum_{t=1}^{\lvert \bm y \rvert} \mathrm{log}\left(p\left(y_t|\bm q,\bm C,\bm S, \bm y_{<t}\right)\right)  \\
        \mathcal L =  & \mathcal L_s + \mathcal L_r
    \end{aligned}
\end{equation}
where $\mathcal L_s$ is the strategy selection loss, $\mathcal L_r$ is the response generation loss, and $L$ is combined objective to minimize.

\section{Experiment}
\subsection{Dataset}
We evaluate our and the compared approaches on ESConv \cite{liu-etal-2021-towards}, which is collected with crowdworkers in a seeker and supporter mode. For preprocessing, we spilt the dataset into train, valid, test with the ratio of 8:1:1. The statistics is given in Table \ref{tab:esconv}. The distribution of strategy is given in Appendix \ref{sec:strategy}.

\begin{table}[htbp]
    \small \centering
    \begin{tabularx}{0.5\textwidth}{lYYY}
        \toprule
        \textbf{Category} & \textbf{Train} & \textbf{Dev} & \textbf{Test} \\ 
        \midrule
        \# dialogues & 1040 & 130 & 130 \\
        \# utterances & 30684 & 3919 & 3762 \\
        Avg. length of dialogue & 29.50 & 30.15 & 28.94 \\
        Avg. length of utterance & 16.50 & 16.13 & 15.81 \\
        \bottomrule
    \end{tabularx}
    \caption{The statistics of processed ESConv dataset.}
    \label{tab:esconv}
\end{table}

\begin{table*}[htbp]
    \small \centering
    \begin{tabularx}{1.0\textwidth}{lcccccccc}
        \toprule
        \textbf{Model} & \textbf{ACC(\%)}$\uparrow$ & \textbf{PPL}$\downarrow$ & \textbf{R-L}$\uparrow$ & \textbf{B-2}$\uparrow$ & \textbf{B-3}$\uparrow$ & \textbf{B-4}$\uparrow$ & \textbf{D-1}$\uparrow$ & \textbf{D-2}$\uparrow$ \\
        \midrule
        Transformer \cite{vaswani2017attention} & - & 114.75 & 14.64 & 6.20 & 3.07 & 1.85 & 0.13 & 0.28 \\
        MT Transformer \cite{rashkin-etal-2019-towards} & - & 109.44 & 14.94 & 5.99 & 2.60 & 1.37 & 0.15 & 0.35 \\
        MoEL \cite{lin-etal-2019-moel} & - & 57.03 & 13.93 & 5.48 & 2.40 & 1.25 & 0.74 & 4.12 \\
        MIME \cite{majumder2020mime} & - & 56.06 & 14.66 & 6.35 & 2.72 & 1.36 & 0.91 & 5.17 \\
        DialoGPT-Joint \cite{zhang-etal-2020-dialogpt} (107M)  & 24.20 & 20.79 & 14.36 & 5.76 & 2.81 & 1.67 & 2.41 & 15.17 \\
        BlenderBot-Joint \cite{roller-etal-2021-recipes} (90M) & 31.17 & 18.60 & 17.03 & 5.80 & 3.09 & 1.88 & 3.08 & 13.95 \\
        DialogVED \cite{chen2022dialogved} (393M) & 22.45 & 19.83 & 16.65 & 5.30 & 3.18 & 2.12 & 4.58 & 16.25 \\
        MultiESC \cite{cheng-etal-2022-improving} (145.6M) & - & \underline{15.41} & \textbf{20.41} & \textbf{9.18} & \textbf{4.99} & \textbf{3.09} & - & - \\
        MISC \cite{tu-etal-2022-misc} (134M) & 31.63 & 16.16 & 17.91 & 7.31 & 3.78 & 2.20 & 4.41 & 19.71 \\
        TransESC \cite{zhao2023transesc} & \textbf{34.71} & 15.85 & 17.51 & 7.64 & 4.01 & 2.43 & \underline{4.73} & \underline{20.48} \\
        ChatGPT (Few-Shot) \cite{OpenAI2022} (175B) & 21.88 & - & 14.03 & 6.10 & 3.00 & 1.68 & \textbf{4.87} & \textbf{22.75} \\ \midrule
        CauESC (Ours) (107M)  & \underline{33.33} & \textbf{15.30} & \underline{18.20} & \underline{8.17} & \underline{4.55} & \underline{2.82} & 4.70 & 19.85 \\ 
        \bottomrule
    \end{tabularx}
    \caption{Automatic Evaluation Results on ESConv. The best results among all models, as well as the second-best results, are highlighted in bold and underlined, respectively.}
    \label{tab:automatic}
\end{table*}

\subsection{Evaluation Metrics}

\paragraph{Automatic Evaluation}
{For the strategy selection, \textbf{ACC} (the accuracy of the strategy) is employed; For the response generation, the conventional \textbf{PPL} (perplexity), \textbf{B-$\bm{n}$} (BLEU-$n$) \cite{papineni-etal-2002-bleu} and \textbf{R-L} (ROUGE-L) \cite{lin-2004-rouge} are utilized as our automatic metrics to evaluate the lexical and semantic aspects of the generation; For response diversity, we report \textbf{D-}$\bm{n}$ (Distinct-$n$) \cite{li-etal-2016-diversity} which evaluates the ratios of the unique n-grams in the generated responses.}

\paragraph{Human Evaluation}
{To ensure consistent evaluations, we utilize human A/B evaluations, which result in a high inter-annotator agreement. Given two models A and B, three annotators are asked to choose the better response for each of the 100 sub-sampled test instances. For objectivity, annotators include those with and without background knowledge (task-related). The final results are determined by majority voting. In case three annotators have three different conclusions, the fourth annotator will bring in. Following \cite{liu-etal-2021-towards}, the aspects contain Fluency, Identification, Comforting, Suggestion and Overall.}

\section{Results and Analyses}
\subsection{Automatic Evaluations}
Table \ref{tab:automatic} displays the performance comparison of different models averaged over three runs.The vanilla Transformer exhibits the poorest performance, as it lacks a specific optimization objective for learning emotional support abilities. Models with empathetic objectives, such as MT Transformer, MoEL, and MIME, also underperform. Their reliance on conversation-level static emotion labels limits their fine-grained emotion understanding.

In comparison to DialoGPT-Joint, BlenderBot-Joint, and DialogVED, CauESC proves more effective in both strategy selection and response generation. While the former models select a strategy at the first decoding step, CauESC utilizes the independent-integrated strategy executors to understand the semantic information of the strategy independently and combines multiple strategies for response generation.

MultiESC outperforms CauESC in R-L, B-2, B-3, and B-4 but lags in PPL. The difference arises from extensive preprocessing in MultiESC on the ESConv dataset, involving manual re-annotation of strategy comments and data augmentation in training (see MultiESC Appendix). These steps may introduce additional manual information, raising fairness concerns. Additionally, MultiESC focuses on strategic planning, diverging from our goal of leveraging emotional causality relationships effectively.

TransESC demonstrates superior performance in ACC, Distinct-1, and Distinct-2 when compared to CauESC. However, it exhibits weaker results in terms of PPL, BLEU-2, BLEU-3, BLEU-4, and ROUGE-L. In its preprocessing phase, TransESC utilizes TF-IDF and fine-tuned BERT for sentiment annotation, raising concerns about fairness. While optimizing for semantic, strategic, and emotional transformation objectives enhances TransESC's accuracy and response diversity slightly over CauESC, it falls short in effectively addressing the interplay between intra-effect and inter-effect. This limitation hampers its performance in language fluency and sentence structure.

In a fair comparison within the experimental setup, CauESC outperforms the state-of-the-art MISC across all metrics, despite having fewer parameters (107M vs. 134M). This suggests that considering strategy history benefits strategy selection. For generation, while MISC incorporates common sense for mental state modeling, CauESC considers emotion causes and adopts emotional dynamics to hierarchically capture emotion effects. It infers emotion effects triggered by emotion causes, providing a fine-grained emotion understanding from cause to effect.

In comparison to ChatGPT, CauESC excels in emotional support conversations. While ChatGPT, with its extensive parameters and reinforcement learning from human feedback \cite{ouyang2022training}, demonstrates proficiency in quickly understanding emotional support requirements and adapting flexibly to few-shot scenarios, it falls short in strategy selection and lacks supportive responses. ChatGPT tends to lean towards \texttt{Providing Suggestions}, offering direct advice to seekers, including overused phrases like “Have you considered,” but often neglects comforting them. This is because ChatGPT is primarily designed for providing information to solve practical problems, which proves insufficient for Emotional Support Conversations (ESC). On the other hand, CauESC accurately selects strategies, incorporating vital information for emotional support tailored to specific seeker situations.

\subsection{Human Evaluations}
In Table \ref{tab:human}, the human evaluations align with automatic evaluations, and for a competitive  and fair comparison, the SOTA model MISC is chosen as a reference. Clearly, CauESC outperforms MISC in all five aspects. Particularly noteworthy is the significant improvement in the Identification metric, indicating CauESC's ability to identify the seeker's dilemma and recognize the emotion causes of distress. In the Comforting metric, CauESC exhibits advancements over MISC, showcasing its ability to capture emotion effects triggered by emotion causes and generate supportive responses accordingly. Overall, responses from CauESC are consistently preferred across all aspects, underscoring the advantages of recognizing emotion causes and reasoning about the effects triggered by those causes.


\begin{table}
    \small \centering
    \begin{tabularx}{0.45\textwidth}{lYYY}
        \toprule
        \textbf{Aspect} & \textbf{Win} & \textbf{Lose} & \textbf{Tie} \\
        \midrule
        Fluency & $\textbf{36}^\ddagger$ & 4 & 60 \\
        Identification & $\textbf{37}^\ddagger$ & 13 & 50 \\
        Comforting & $\textbf{62}^\ddagger$ & 6 & 32 \\
        Suggestion & $\textbf{24}^{\dagger}$ & 12 & 64 \\
        Overall & $\textbf{54}^\ddagger$ & 17 & 29 \\
        \bottomrule
    \end{tabularx}
    \caption{Human Evaluation Results on ESConv. $\dagger$ and $\ddagger$ represent improvement with $p$-value < 0.1/0.05, respectively. Tie indicates responses from both models are deemed equal. The Fleiss Kappa score \cite{fleiss1973equivalence} reaches 0.406, indicating a moderate level of agreement.}
    \label{tab:human}
\end{table}

\subsection{Ablation Study}
To verify the improvement brought by each added component, the ablation study is performed in Table \ref{tab:ablation}. Results have shown that each component is beneficial to the final result. (1) Emotion causes contribute toward CauESC’s all performance, indicating that recognizing the emotion causes is necessary to provide emotional support. (2) There is some decrease when intra or inter effects are removed, suggesting that capturing emotional dynamics in conversations is beneficial to understand the seeker's mental state. (3) Notably, removing strategy executors leads to a deep drop of B-2 and B-4, which demonstrates that the effectiveness of strategy executors, which allows the model to independently understand the semantic information of each strategy. (4) Although some components are removed, the quality of CauESC's responses are still better than BlenderBot-Joint \cite{liu-etal-2021-towards}, demonstrating the stability of CauESC.

\begin{table}
    \small \centering
    \begin{tabularx}{0.45\textwidth}{lYYYY}
        \toprule
        \textbf{Model} & \textbf{R-L}$\uparrow$ & \textbf{B-2}$\uparrow$ & \textbf{B-4}$\uparrow$ & \textbf{D-1}$\uparrow$ \\
        \midrule
        CauESC & \textbf{18.20} & \textbf{8.17} & \textbf{2.82} & \textbf{4.70} \\
        \midrule
        w/o Cause & 18.06 & 7.92 & 2.67 & 4.54 \\
        w/o Intra Effect & 18.13 & 7.82 & 2.58 & 4.49 \\
        w/o Inter Effect & 17.79 & 8.16 & 2.66 & 4.05 \\
        w/o Executors & 18.12 & 7.64 & 2.56 & 4.37 \\
        \midrule
        BlenderBot-Joint & 17.03 & 5.80 & 1.88 & 3.08 \\
        \bottomrule
    \end{tabularx}
    \caption{Evaluation Results of Ablation Study.}
    \label{tab:ablation}
\end{table}

\begin{table*}[htbp]
    \small \centering
    \begin{tabularx}{\textwidth}{l|Z}
        \Xhline{.08em}  
        \multicolumn{2}{c}{\textbf{Situation}} \\
        \hline
        Seeker & My girl decided to break up with me by text, instead of in person. Other friends also heard... \\
        \hline
        \multicolumn{2}{c}{\textbf{Context}} \\
        \hline
        Seeker & \textcolor[RGB]{154,3,30}{My friends came up to me in the bar and said they had heard about the breakup...} \textcolor[RGB]{0,119,182}{\{\texttt{xReact:} upset, \texttt{xWant:} leave\}} \\
        \hline
        Supporter & Yes it is a very cowardly thing to do. I am quite disgusted that she did not have the courage to talk to you... \textcolor[RGB]{0,119,182}{\{\texttt{oEffect:} to be safe, \texttt{oWant:} tell person x\}} \\
        \hline
        Seeker & \textcolor[RGB]{154,3,30}{Some friends to gossip...yeah, I am disgusted for sure. We had been together over 2 years...} \textcolor[RGB]{0,119,182}{\{\texttt{xEffect:} get angry, \texttt{xWant:} to continue in relationship\}}\\
        \hline
        \multicolumn{2}{c}{\textbf{Response}} \\
        \hline
        \textbf{Ground-Truth} & Unfortunately there are people out there who are not true friends and who just want something to talk about. I have experienced that myself. \\
        \hline
        BlenderBot-Joint & I know how you feel. I have been in a similar situation. (\textit{General Response}) \\
        \hline
        MISC & Do you think you should talk to your friends about it? (\textit{Poor Empathy})  \\
        \hline
        ChatGPT & I can understand why you feel that way. It's completely normal to feel disgusted and hurt when someone you love breaks up with you in such a disrespectful way. (\textit{Misunderstanding of Emotion Cause}) \\ 
        \hline
        CauESC & I would be disgusted too. I have been in a similar situation myself. It is difficult to open up to someone who does not respect you. (\textit{Specific and Strong Support}) \\
        \Xhline{.08em}
    \end{tabularx}
    \caption{Case study of the generated responses by baselines and CauESC. \textcolor[RGB]{154,3,30}{Red} represents the emotion causes of the seeker. \textcolor[RGB]{0,119,182}{Blue} indicates the emotion effects generated by COMET. \textit{Italic} represents a simple comment of the response. Some words are omitted due to space limit.}
    \label{tab:case_study}
\end{table*}

\subsection{Strategy Modeling Analysis}
To assess the benefit of the independent-integrated strategy modeling approach quantitatively, we implement different strategy modeling approaches based on CauESC: (1) \textbf{CauESC-Label:} The executors no longer learn the description of strategy, but a single strategy label. (2) \textbf{CauESC-Multi:} Following the mixed strategy modeling of MISC \cite{tu-etal-2022-misc}, we employ a multi-factor-aware decoder for mixed strategy modeling instead of independent-integrated strategy executors. (3) \textbf{CauESC-Single:} Following the single strategy modeling of BlenderBot-Joint \cite{liu-etal-2021-towards}, a single strategy token is generated at the first decoding step. The comparison results are given in Table \ref{tab:strategy_modeling}.

\begin{table}[htbp]
    \small \centering
    \begin{tabularx}{0.45\textwidth}{lYYYY}
        \toprule
        \textbf{Model} & \textbf{R-L}$\uparrow$ & \textbf{B-2}$\uparrow$ & \textbf{B-4}$\uparrow$ & \textbf{D-1}$\uparrow$ \\
        \midrule
        CauESC & \textbf{18.20} & \textbf{8.17} & \textbf{2.82} & \textbf{4.70} \\ \midrule
        CauESC-Label & 18.11 & 8.06 & 2.69 & 4.27 \\
        CauESC-Multi & 18.05 & 7.97 & 2.51 & 4.10 \\
        CauESC-Single & 18.12 & 7.64 & 2.56 & 4.37 \\
        \bottomrule
    \end{tabularx}
    \caption{Comparison Results of Strategy Modeling.}
    \label{tab:strategy_modeling}
\end{table}

The performance of CauESC-Label decreases, indicating the necessity of comprehending the meaning of the strategy for generating emotional support. CauESC-Multi fares notably worse, emphasizing the efficacy of CauESC's independent-integrated strategy modeling. This approach enables independent grasping of each strategy's semantic information, in contrast to MISC's mixed strategy modeling that neglects their independence. Finally, CauESC-Single exhibits significant degradation, particularly in B-2 and B-4, underscoring the advantage of integrating strategies to guide response generation. In summary, independent-integrated strategy modeling proves more effective than other strategy modeling approaches.

\subsection{Case Study}
As shown in Table \ref{tab:case_study}, an example is present to compare responses generated by CauESC and other baselines. Problems appear in the baselines, such as general response and misunderstanding of emotion cause. Intuitively, our model achieves the best performance and shows many highlights.

\noindent\textbf{Hightlight 1: CauESC effectively recognizes the emotion causes of the seeker and reasons about the emotion effects.} Specifically, CauESC is successful in recognizing \deleted{the }the emotion cause which is the gossip about the breakup spread by friends. However, by comparison, ChatGPT incorrectly assumes the distress is caused by the breakup, which generates unrelated and unresonant response. Moreover, CauESC also captures the emotion effects triggered by the cause, i.e., disgust with friends, with related COMET commonsense \{\texttt{xReact:} upset\} and \{\texttt{xEffect:} get angry\}, leading to the specific and supportive response \textit{“it is difficult to open up to someone who does not respect you”} which reflects the fine-grained emotion understanding from cause to effect.

\noindent\textbf{Hightlight 2: CauESC understands each strategy independently and integrates them skillfully.} After capturing the seeker’s emotion effects, CauESC firstly decides to employ the strategy of \texttt{Reflection of Feelings} to emotionally express its understanding of the disgusting feelings. Then, it indicates that it had similar situation with the strategy of \texttt{Self-disclosure}. Finally, it also supplements detailed information according to the emotion causes and effects to suggest that friends are not worth cherishing which could be regarded as using the strategy of \texttt{Information}.

\section{Conclusions}
In this paper, we propose CauESC, a causal aware model for emotional support conversation, which recognizes the emotion causes, captures the emotion effects triggered by the causes, and adopt an independent-integrated strategy modeling approach. Through extensive experiments and both automatic and human evaluations, we have demonstrated the advantages of CauESC in emotional support conversation system.


\section*{Limitations}
Although we confirm the effectiveness of our proposed framework through extensive experiments, the research on emotional support conversation still faces significant challenges and opportunities for further advancement lie ahead. Compared with the human supporters, while the chatbots demonstrate the capability to comfort the seeker, it struggles to offer constructive advice on how to solve the problems in the real situation. This problem may be alleviated by introducing knowledge graphs. We can retrieve knowledge from knowledge graph about how to tackle real-life issues and incorporate relevant knowledge into the response to provide specific and constructive advice to the seeker, which will be included in our future research. Other issues, such as how to select support strategies more accurately and how to generate personalized responses in few-shot and low-resource scenarios, also need to be further explored.

\section*{Ethics Statement}
The ethical considerations surrounding our work are of utmost importance to us. In this section, we discuss the potential ethic impacts of this work: (1) We utilize the ESConv dataset, which is a publicly available and well-established benchmark for emotional support conversation. By using this dataset, we ensure transparency and reproducibility in our experiments, allowing other researchers to evaluate and build upon our work. (2) We prioritize the protection of privacy. \citet{liu-etal-2021-towards} have taken measures to filter sensitive information on ESConv. Note that the filtering process may not cover all instances of emotionally triggering language. (3) We emphasize that our aim is to provide emotional support and facilitate emotional support conversations, while acknowledging the importance of referring individuals to appropriate professional help in such situations. In cases involving risky situations, we do not claim to provide any form of treatment or diagnosis.



\appendix

\section{Experimental Setup}
\paragraph{Compared Models}
{Our baselines include a vanilla \textbf{Transformer} \cite{vaswani2017attention}; three empathetic chatbots \textbf{MT Transformer} \cite{rashkin-etal-2019-towards}, \textbf{MoEL} \cite{lin-etal-2019-moel}, and \textbf{MIME} \cite{majumder2020mime}; three SOTA methods on the ESConv \textbf{DialoGPT-Joint} \cite{zhang-etal-2020-dialogpt}, \textbf{BlenderBot-Joint} \cite{roller-etal-2021-recipes}, and \textbf{MISC} \cite{tu-etal-2022-misc}; a pre-trained latent variable encoder-decoder model \textbf{DialogVED} \cite{chen2022dialogved}; and a large language model \textbf{ChatGPT} \cite{OpenAI2022} with few-shot prompting. More details are described in the Appendix \ref{sec:models}, and details about few-shot prompting of ChatGPT are shown in the Appendix \ref{sec:chatgpt}.}

\paragraph{Implementation Details}
{We implement our approach based on blenderbot-small \cite{roller-etal-2021-recipes} with Transformers \cite{wolf-etal-2020-transformers}. To annotate emotion causes on ESConv, we use RoBERTa-Large \cite {Liu2019RoBERTaAR} fine-tuned on the RECCON \cite{poria2021recognizing} as an emotion cause detector. To be comparable with the SOTA model in \cite{tu-etal-2022-misc}, we fine-tune CauESC based on the blenderbot-small with 90M parameters by a RTX-3090 GPU. The batch size of training and evaluating is 20 and 50, respectively. We set the epoch to 8, initialize the learning rate as 2e-5 and change it using a linear warmup with 120 warmup steps. We use AdamW as optimizer \cite{Loshchilov2017DecoupledWD} with $\beta_1=0.9$ , $\beta_2=0.999$ , $\epsilon=\text{1e-8}$. Following \citet{tu-etal-2022-misc}, we also adopt the decoding algorithms of Top-$p$ and Top-$k$ sampling with $p=0.3$ , $k=30$ , temperature $\tau=0.7$ and the repetition penalty 1.03.  The source code will be released to facilitate future work.}

\section{Definition of COMET Relations}
\label{sec:comet}
\noindent\texttt{oEffect} The effect the event has on others besides Person X.

\noindent\texttt{oReact} The reaction of others besides Person X to the event.

\noindent\texttt{oWant} What others besides Person X may want to do after the event.

\noindent\texttt{xAttr} How Person X might be described given their part in the event.

\noindent\texttt{xEffect} The effect that the event would have on Person X.

\noindent\texttt{xIntent} The reason why X would cause the event.

\noindent\texttt{xNeed} What Person X might need to do before the event.

\noindent\texttt{xReact} The reaction that Person X would have to the event.

\noindent\texttt{xWant} What Person X may want to do after the event.

\section{Description of Strategy Reference}
\label{sec:strategy_reference}
Following \citet{liu-etal-2021-towards}, the description of each strategy is defined as follows.

\noindent\texttt{Question} Asking for information related to the problem to help the help-seeker articulate the issues that they face. Openended questions are best, and closed questions can be used to get specific information.

\noindent\texttt{Restatement or Paraphrasing} A simple, more concise rephrasing of the help-seeker’s statements that could help them see their situation more clearly.

\noindent\texttt{Reflection of Feelings} Articulate and describe the helpseeker’s feelings.

\noindent\texttt{Self-disclosure} Divulge similar experiences that you have had or emotions that you share with the help-seeker to express your empathy.

\noindent\texttt{Affirmation and Reassurance} Affirm the help-seeker’s strengths, motivation, and capabilities and provide reassurance and encouragement.

\noindent\texttt{Providing Suggestions} Provide suggestions about how to change the situation, but be careful to not overstep and tell them what to do.

\noindent\texttt{Information} Provide useful information to the help-seeker, for example with data, facts, opinions, resources, or by answering questions.

\noindent\texttt{Others} Exchange pleasantries and use other support strategies that do not fall into the above categories.

\section{Distribution of Strategy}
\label{sec:strategy}
\begin{figure}[htbp]
  \centering
  \includegraphics[width=0.5\textwidth]{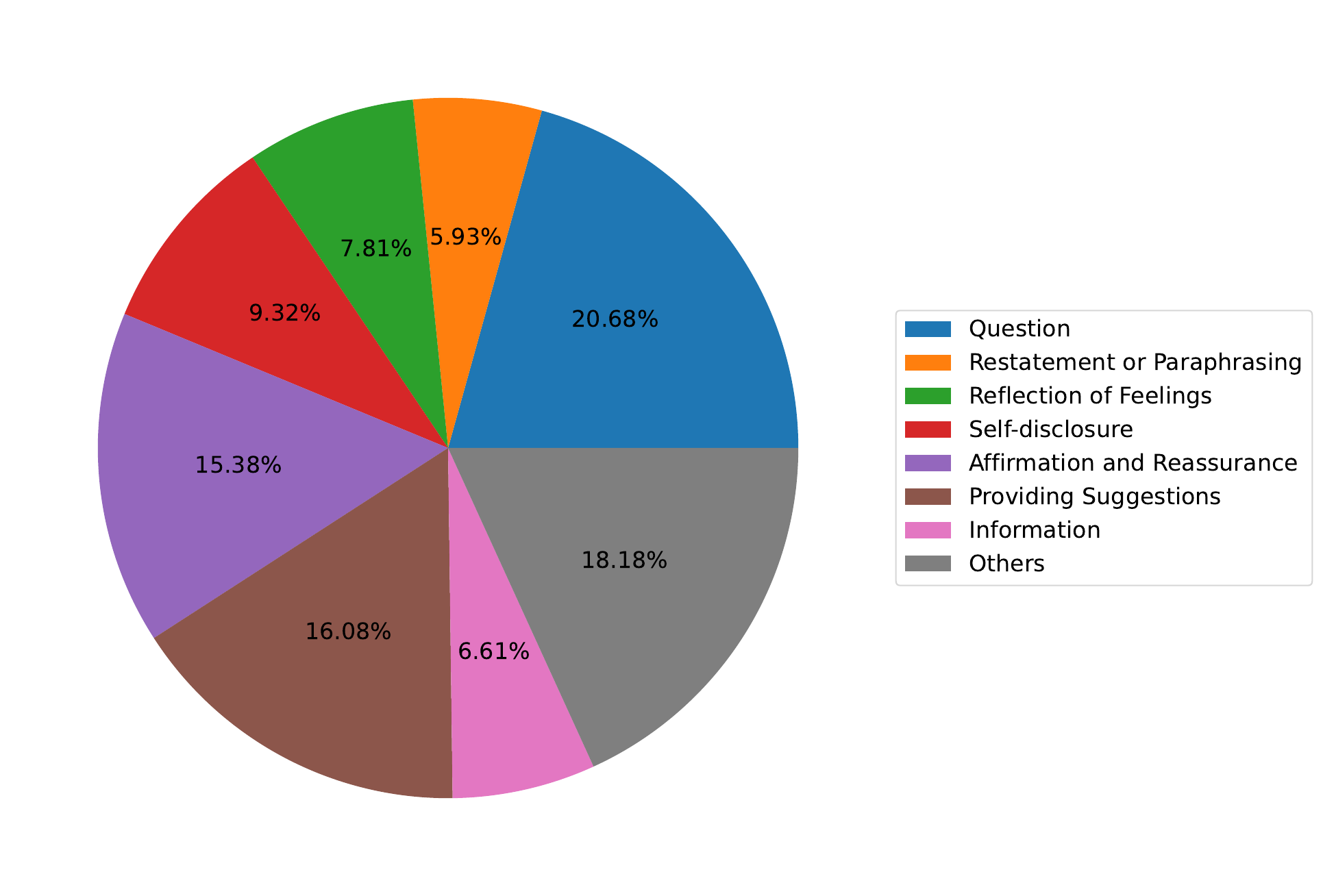}
  \caption{The strategy distribution on ESConv.}
  \label{fig:strategy}
\end{figure}

As shown in Figure \ref{fig:strategy}, we counted the strategy distribution on ESConv. It can be observed that the proportion of each strategy is relatively balanced, which allows the model to learn each strategy comprehensively.

\begin{figure}[htbp]
  \centering
  \includegraphics[width=0.5\textwidth]{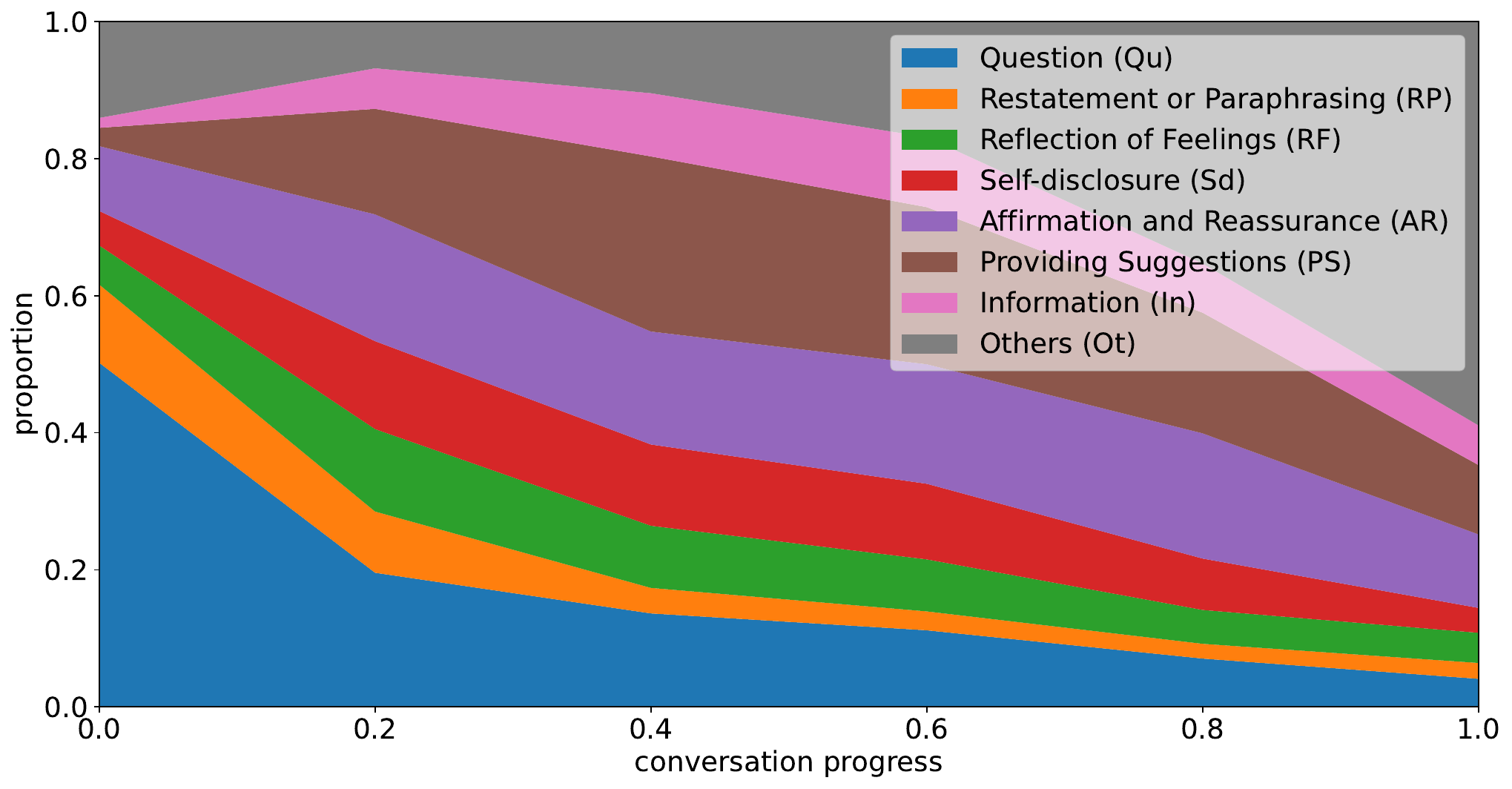}
  \caption{The strategy distribution at different conversation progress on ESConv.}
  \label{fig:strategy_progress}
\end{figure}

Furthermore, as shown in Figure \ref{fig:strategy_progress}, we computed the distribution of strategies at different phases of the conversation. Specifically, We split the conversation progress into six intervals and calculated the distribution of strategies for each interval in chronological order respectively and connected them. It can be seen that the distribution of strategies changes as the conversation progresses. For example, in the early stages of the conversation, supporters usually adopt exploratory strategies, such as \texttt{Question}. After learning about the emotion causes of the seeker's distress, the supporter tends to offer their opinion  (such as \texttt{Providing Suggestions}). Reassuring strategies (such as \texttt{Affirmation and Reassurance}) are used throughout the conversation and label a relatively constant proportion of messages.

\section{Compared Models}
\label{sec:models}
\noindent\textbf{Transformer} A vanilla Seq2Seq model includes an encoder and a decoder \cite{vaswani2017attention}.

\noindent\textbf{MT Transformer} A variation of the Transformer that has an additional task for predicting the emotion \cite{rashkin-etal-2019-towards}.

\noindent\textbf{MoEL} A Transformer-based model combines representations from multiple decoders to improve the empathy \cite{lin-etal-2019-moel}.

\noindent\textbf{MIME} Another Transformerbased model which
leverages the emotional polarity and mimicry for empathetic response generation \cite{majumder2020mime}.

\noindent\textbf{DialoGPT-Joint} An open-domain conversational agent, which appends a special strategy token before the response utterances to make a generation \cite{zhang-etal-2020-dialogpt}.

\noindent\textbf{BlenderBot-Joint} Another chatbot trained on conversation skill datasets which also appends a strategy token before generation \cite{roller-etal-2021-recipes}.

\noindent\textbf{DialogVED} A pre-trained dialogue system that introduces latent variables and achieves good performance on multiple downstream dialogue tasks with the same strategy modeling approach as Blenderbot-Joint \cite{chen2022dialogved}.

\noindent\textbf{MISC} A mixed strategy-aware model integrating COMET to capture seeker’s mental state, and generate responses with the mixed strategies, which is the SOTA model on ESConv \cite{tu-etal-2022-misc}.

\noindent\textbf{ChatGPT} A large language model fine-tuned from the GPT-3.5 series, which is trained to follow an instruction in a prompt and provide a detailed response, using Reinforcement Learning from Human Feedback (RLHF) \cite{OpenAI2022}. In this paper, we use few-shot prompting to allow the model to perform in-context learning, rather than fine-tuning. For strategy modeling, It uses the same strategic modeling approach as BlenderBot-Joint.

For reference, DialoGPT-Joint, BlenderBot-Joint, DialoVED, MISC and ChatGPT have 90M, 117M, 393M, 134M, 175B parameters respectively, and our model CauESC has 107M parameters which is in the same order as the baselines except ChatGPT.

\section{Prompt of ChatGPT}
\label{sec:chatgpt}
\begin{figure}[htbp]
  \centering
  \includegraphics[width=0.5\textwidth]{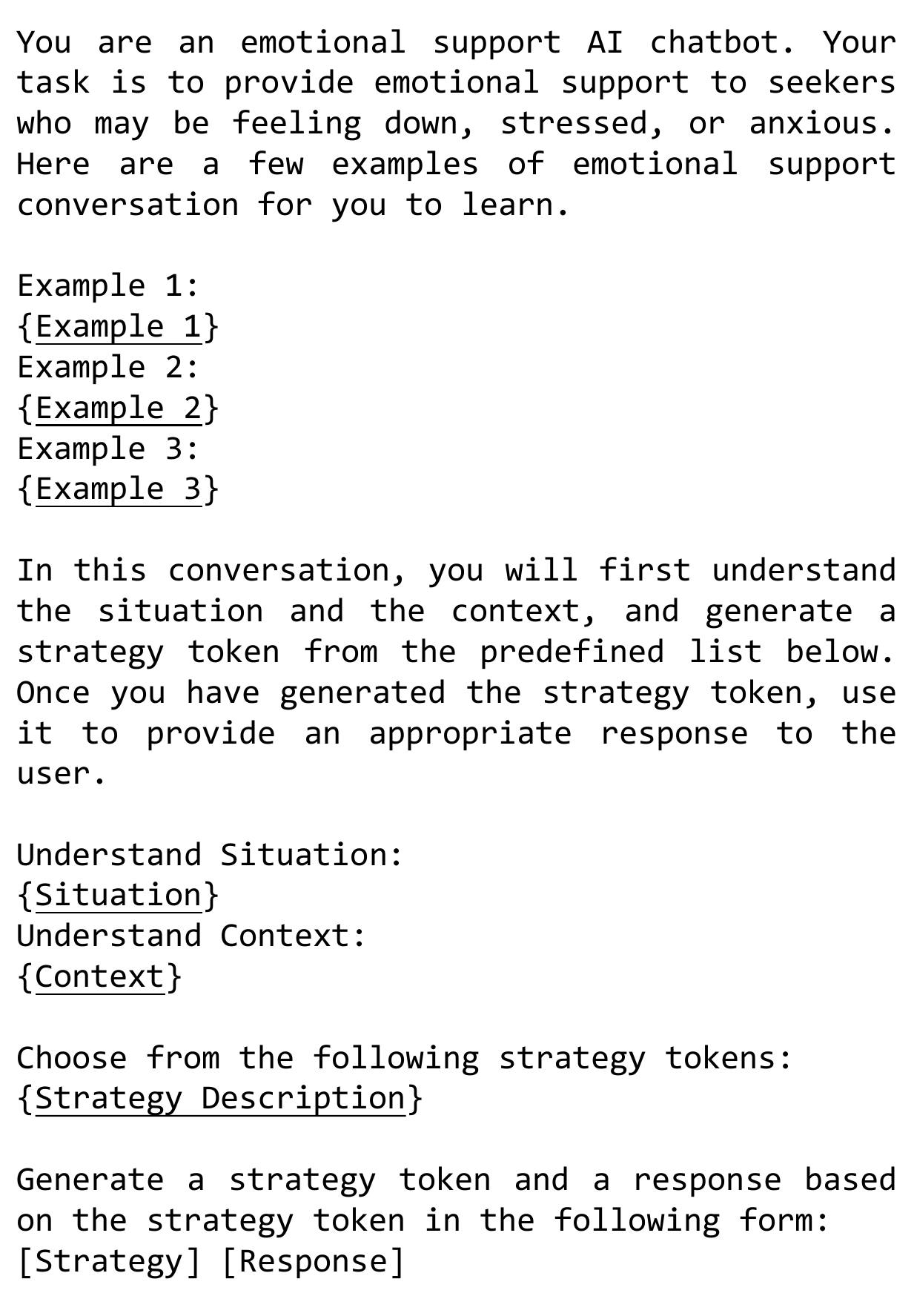}
  \caption{The few-shot prompt of ChatGPT.}
  \label{fig:prompt}
\end{figure}

In order for ChatGPT to understand the task requirements of emotional support conversation, we design the following few-shot prompt including task descriptions, examples, guidance, situation, context, strategy description, and formatting requirements, as shown in Figure \ref{fig:prompt}.

In the prompt, \texttt{\{\underline{Example}\}} represents an emotional support conversation example including the situation and the context; \texttt{\{\underline{Situation}\}} indicates the seeker's situation of the current conversation; \texttt{\{\underline{Context}\}} denotes the context of the current conversation; and \texttt{\{\underline{Strategy Description}\}} contains the strategies and their descriptions in the same way as Strategy Reference shown in Appendix \ref{sec:strategy_reference}.

For setup details of ChatGPT, we use Chat Completions API for generation, where chat models take a list of messages as input and return a generated message as output. Although the chat format is designed to make multi-turn conversations easy, it’s just as useful for single-turn tasks without any conversation. The conversation is formatted with a system message first, which is the prompt above to set the behavior of the assistant. As for parameters, we set gpt-3.5-turbo as ID of the model. For a fair comparison, we also adopt the decoding algorithms of Top-$p$ sampling with $p=0.3$ , temperature $\tau=0.7$ and the presence penalty 1.03.

\end{document}